\pdfoutput=1
\documentclass[11pt]{article}

\usepackage[preprint]{acl}

\usepackage{times}
\usepackage{latexsym}
\usepackage{amsmath, amssymb}
\usepackage[T1]{fontenc}
\usepackage[utf8]{inputenc}

\usepackage{microtype}

\usepackage{inconsolata}

\usepackage{graphicx}
\usepackage{booktabs} 
\title{Decipherment-Aware Multilingual Learning in Jointly Trained Language Models}

\author{Grandee Lee \\
  Singapore University of Social Sciences\\
  grandeelee@suss.edu.sg \\ \\
  \texttt{Mar, 2021}\\}

\begin{document}
\maketitle
\begin{abstract}
The principle that governs unsupervised multilingual learning (UCL)\footnote{\textit{Multilingual learning} is interchangeable with \textit{cross-lingal learning}, we keep to unsupervised cross-lingual learning (UCL) following the literature.} in jointly trained language models (mBERT as a popular example) is still being debated. Many find it surprising that one can achieve UCL with multiple monolingual corpora.
In this work, we anchor UCL in the context of language decipherment and show that the joint training methodology is a decipherment process pivotal for UCL. In a controlled setting, we investigate the effect of different decipherment settings on the multilingual learning performance and consolidate the existing opinions on the contributing factors to multilinguality. From an information-theoretic perspective we draw a limit to the UCL performance and demonstrate the importance of token alignment in challenging decipherment settings caused by differences in the data domain, language order and tokenization granularity. Lastly, we apply lexical alignment to mBERT and investigate the contribution of aligning different lexicon groups to downstream performance. 
\end{abstract}

\section{Introduction}
Multilingual learning models aim to connect and empower the innumerable languages around the world. We refer to \citet{artetxe-etal-2020-call}'s definition of  \textit{``Multilingual learning'' as learning a common model for two or more languages from raw text, without any downstream task labels}. The term is interchangeable with \textit{``cross-lingual learning''}. 
This pretrained model can be fine-tuned for various downstream tasks, e.g.\ cross-lingual classification~\cite{conneau-etal-2018-xnli, yang-etal-2019-paws}, structural prediction~\cite{nivre-etal-2020-universal, pan-etal-2017-cross} and retrieval~\cite{zweigenbaum-etal-2018-overview, artetxe-schwenk-2019-massively}. This paper focuses on the Unsupervised Cross-lingual Learning (UCL) of jointly trained models, a popular example being mBERT~\cite{devlin-etal-2019-bert}. 

mBERT and similar models, e.g.\ XLM-R~\cite{conneau-lample-2019-cross, conneau-etal-2020-unsupervised} exhibit surprisingly good zero-shot cross-lingual transfer learning for downstream tasks given their unsupervised training methodology. 
These models represent various languages in a shared space that enables the cross-lingual transfer of languages. However, they do so without relying on explicit cross-lingual signals\footnote{cross-lingual signals could be word-level correspondence e.g.\ a bilingual dictionary, sentence-level, e.g.\ bitext or at document-level}. The training methodology is characterized by 1) joint training on a combination of multilingual datasets, and with 2) the masked language modeling (MLM) objective function, that is referred to  as the UCL implementation hereafter. 
Studies have shown that UCL under such training scheme  depends neither on the shared vocabulary nor on possible anchor words present in multiple datasets~\cite{pires-etal-2019-multilingual, wu-dredze-2020-explicit, K-2020-cross-Lingual, conneau-etal-2020-emerging}. Therefore, the question remains as to what governs UCL in jointly trained models. 

We would like to offer a perspective by connecting the UCL training methodology to the decipherment task, with the aim of improving this category of models in mind.
However, our investigation departs from the previous studies that largely focus on the internal mechanisms of the model architecture (\S\ref{sec:related_work}). 
In \S\ref{sec:decipher}, we define multilingual learning based on the decipherment task and devise 9 bilingual decipherment settings of varying distributional divergence. In \S\ref{sec:eval_setup}, we propose a set of evaluation metrics for both UCL and decipherment performance.
We show that the UCL is highly correlated to the decipherment performance and decipherment difficulty, which in turn are an interplay of the data domain, language order, lexical granularity of the input datasets, as well as the language modeling objective, and joint training (\S\ref{sec:analysis}). 
Lastly by improving the decipherment performance, we are able to improve mBERT's cross-lingual performance on an array of downstream tasks (\S\ref{sec:expt_setup}).

\section{Background}
\label{sec:related_work}
In this section, we briefly review relevant work on UCL from Cross-Lingual Word Embedding (CLWE) to, more recently, jointly trained models, as well as the work related to language decipherment.

\paragraph{CLWE}
Both CLWE and deep multilingual models share the same goal of deriving better (sub)word representations. Their representation differ w.r.t.\ whether it is static or contextual~\cite{artetxe-etal-2020-call}. In CLWE, cross-lingual signals can be gradually reduced if we make an increasingly stronger assumption on the \textit{``structure''} similarity of the two languages. From relying on dictionary~\cite{mikolov-etal-2013-exploiting, faruqui-dyer-2014-improving, huang-etal-2015-translation, zhang-etal-2016-ten, 8683678} to relying on similar documents~\cite{Vulic:2016, levy-etal-2017-strong, lee19d_interspeech}. Ultimately in the case of unsupervised CLWE, we can rely on the isomorphic structure~\cite{lample-etal-2018-word, artetxe-etal-2018-robust, zhang-etal-2017-adversarial, lee-li-2020-modeling} of the languages. One may refer to \citet{Ruder_2019} for a more comprehensive survey. Knowing that CLWE has to rely on the distributions of the languages as a form of cross-lingual signal, likewise a jointly trained model such as mBERT. However, it is not apparent what signals are utilized and how joint learning with a masking objective can utilize them. We will provide explanations to the effectiveness of such jointly trained models from a language decipherment perspective.

\paragraph{Jointly trained models}
Moving to deep contextual multilingual models, such as mBERT~\cite{devlin-etal-2019-bert}, XLM-R~\cite{conneau-etal-2020-unsupervised}, there seems to be an absence of cross-lingual signal and explicit cross-lingual assumptions, except that the models are jointly trained. 
\citet{pires-etal-2019-multilingual} and \citet{K-2020-cross-Lingual} observe that mBERT is cross-lingually effective even without any lexical overlap, i.e., the embedding spaces w.r.t different languages can be disjoint. The observation is supported by \citet{conneau-etal-2020-emerging}, which further points out that the crucial step in multilingual learning is the shared layers, not the shared vocabulary. In the same study, canonical correlation analysis reveals a high degree of association between the layer activations in monolingual transformer models. We hypothesize that mBERT relies on this high symmetry in the monolingual model structures to align the disjoint embedding spaces implicitly. Importantly, we identify joint training as a form of language decipherment.

\citet{conneau-etal-2020-emerging} note the effect of different topological features and observe that cross-lingual transfer is less effective in distant language pairs. \citet{wu-dredze-2020-languages} look into the difference in data size and the bias introduced by subword encoding. \citet{dufter-schutze-2020-identifying} also identify the various architectural and linguistic factors contributing to UCL. Our study focuses on word distribution of the input corpora and the effect of decipherment difficulty on UCL.

\paragraph{Decipherment}
Statistical decipherment~\cite{ravi-knight-2008-attacking, luo-etal-2021-deciphering} can trace to the early work by \citet{shannon-1949-communication} who defined the limit of entropy of decoding the target text based on the source text length. More relevant to this work is the application to bilingual lexicon induction (BLI). Learning from non-parallel data, BLI has a rich history relying on statistical analysis and decipherment methods~\cite{rapp-1995-identifying, fung-1995-compiling, koehn-knight-2002-learning, haghighi-etal-2008-learning}. 
We extend the information-theoretic analysis to the case of UCL.

In a related setting of machine translation, \citet{ravi-knight-2011-deciphering, nuhn-etal-2012-deciphering, dou-knight-2013-dependency} use probabilistic language models to decipher languages without parallel data. In a similar vein but using neural approach,~\citet{lample-etal-2018-unsupervised} and~\citet{artetxe-etal-2018-unsupervised} trained translation models with only the respective language corpora and a pre-aligned embedding space. We draw inspirations from these lines of works and develop the theory of neural decipherment based on joint training.

\section{Deciperment Perspective on Joint Training}
\label{sec:decipher}
Let's consider the current UCL paradigm as the composite of two distinct processes. Firstly, the sharing of word spaces, i.e.\ at the lexical level similar to CLWE, and secondly, the learning of contextual representations enabled by deep models. 
This view is reminiscent of the works in CLWE whereby multilingual learning can be tackled by establishing the common words space first and then building up the contextual representation for downstream tasks~\cite{lample-etal-2018-unsupervised, artetxe-etal-2020-cross}. 
While there is a line of work studying the contextual representation of deep language models, it is not obvious why joint training, without explicit cross-lingual signals, enables the sharing of word embedding spaces.

Our discussions here focus on two languages and the problem is similar to BLI from non-parallel corpora. We take a decipherment perspective \S\ref{sec:neural_decipher} and link to mBERT in \S\ref{sec:mbert_decipher}. 

\subsection{Neural Decipherment}
\label{sec:neural_decipher}
In a decipherment task we have \textit{ciphertext} as $f_1^N = f_1 \dots f_j \dots f_N$ consisting of cipher tokens\footnote{We use tokens to denote both words and subwords.} $f_j \in V_f$. We also have \textit{plaintext} as $e_1^M = e_1 \dots e_i \dots e_M$ with $e_i \in V_e$. They are interchangeably referred to as the target and source $(f, e)$ languages. 
In a substitution cipher, a given plaintext is encrypted into ciphertext by replacing each plaintext token with a unique cipher token. 

The objective is to learn a bijective function $\phi: V_f \rightarrow V_e$ that maps the tokens in $f$ to the tokens in $e$, i.e.\ the character or word level correspondence between the two texts. 
This is done by finding $\phi$ that maximizes the fluency of the deciphered text $\phi(f_1^N)$ (\ref{equ:decipher}). The fluency is scored using a language model, $p(\cdot)$, trained on the plaintext corpus. 
\begin{equation}
\label{equ:decipher}
    \hat{\phi} = \arg\max_{\phi} p(\phi(f_1)\phi(f_2)\phi(f_3)\dots\phi(f_N))
\end{equation}

Unlike the statistical decipherment approaches to word translation problem~\cite{nuhn-etal-2013-beam, ravi-knight-2011-deciphering, luo-etal-2019-neural},  a neural language model directly derives a mapping $\phi$ from the word embedding space. 

\paragraph{Solving a substitution cipher}
Given two corpora, $D_{e}, D_{f}$, we first optimize a language model on the source language, i.e. \textit{plaintext}, $e$ according to Eq. (\ref{equ:lm1}). We explicitly treat the contextual layers $\mathbf{W}_e$ and the word embedding $\mathbf{E}_e$ as two distinct parts of the language model. In the decipherment step (\ref{equ:lm2}), we freeze $\mathbf{W}_e$ and retrain the model on the target language, $D_{f}$, to learn a new $\mathbf{E}_{f}$. 
\begin{align}
    \label{equ:lm1}
    \mathcal{L}_{e}(\mathbf{W}_{e}, \mathbf{E}_{e}) &= \mathbb{E}_{t_e{\raise.12ex\hbox{$\scriptstyle\sim$}} D_{e}}[p(t_e) ] \\
    \label{equ:lm2}
    \mathcal{L}_{f}(\overline{\mathbf{W}}_{e}, \mathbf{E}_{f}, \phi) &= \mathbb{E}_{t_f{\raise.12ex\hbox{$\scriptstyle\sim$}} D_{f}}[ p ({\phi}(t_f)) ] 
\end{align}
where $\phi$ is not explicitly optimized, but rather computed from the cosine distance between $(\mathbf{E}_{e}, \mathbf{E}_{f})$. $\phi$ will recover the optimal mapping for deciphering $f$ because the fixed $\overline{\mathbf{W}}_e$ serves as the constraint in the word embedding optimization and will prefer an $\mathbf{E}_{f}$ that follows the distribution 
of $\mathbf{E}_{e}$. $D_f$ and $D_e$ have to be compatible corpora. 

The neural decipherment method consumes more data to match the statistical methods' performance. But, the $\mathbf{W}, \mathbf{E}$ pair paves the way for joint decipherment.
This setup is also reminiscent of the adversarial learning in CLWE, with $\mathbf{W}_e$ as the discriminator and $\phi(\mathbf{E}_{f})$ as the embedding generator, that maximizes the similarity of the two embedding spaces \cite{miceli-barone-2016-towards, zhang-etal-2017-adversarial}.
It's worth noting that the decipherment method is also a viable method in unsupervised CLWE.
In character level experiments we can achieve perfect decipherment, we can provide more details.

\subsection{Joint Training as Bidirectional Decipherment}
\label{sec:mbert_decipher}

In \S\ref{sec:neural_decipher}, we present the unidirectional decipherment setup, $f \rightarrow e$. The resultant model parameters are $\mathbf{E}_f$, $\mathbf{E}_e$, and $\mathbf{W}_e$ which are biased towards the distribution of $D_e$. 
Under joint training, the decipherment is bidirectional. The model bootstraps from what is learnt in both directions.
In this way, $\mathbf{E}_f$ and $\mathbf{E}_e$ are constrained by the shared contextual layers, $\mathbf{W}_{e \leftrightarrow f}$, during training and simultaneously fulfil the distributional preference of both languages. 
This is more advantageous than a $f \rightarrow e$ decipherment and enables more isomorphic representations \cite{ormazabal-etal-2019-analyzing}. 
This perspective agrees with the findings by \citet{conneau-etal-2020-emerging} that the sharing of parameters $\mathbf{W}_{e \leftrightarrow f}$ is crucial in UCL. 

If UCL is indeed underpinned by the decipherment principle, then naturally the decipherment performance and difficulty will affect the performance of UCL. To validate this hypothesis in a controlled setting, we devise different decipherment setups of varying difficulty levels and measure the UCL performance.

\section{Decipherment Experiments}
\label{sec:eval_setup}
What affects the quality of decipherment includes the equivalency of codes, referred to as tokens hereafter, between the source and target codes.
We identified three factors that affect the token distributions, they are the input domain, lexical granularity and token order.
In order to fully recover the token level correspondence, $\phi$, between the source and target tokens, the distribution of the corresponding token pairs in the two datasets should be similar.
Diverging distribution will present a more challenging decipherment setting. Ultimately, language difference is the main contributor to decipherment difficulty. However, for reasons that will be apparent later, we adopt a controlled setting that only simulates the language difference. The proposed settings are indicative of the actual UCL behaviour between languages.

\subsection{Experiment Setup}
To start with the ideal case of substitution decipherment, we use the English and Fake-English setup \cite{K-2020-cross-Lingual, dufter-schutze-2020-identifying}. The Fake-English assumes the role of the target language, $f$. Both $D_f$ and $D_e$ are identical except that each token indices are shifted by a constant and tokens are prefixed with ``::'' before adding to the vocabulary, following the setup described in \citet{dufter-schutze-2020-identifying}. 
This setup draws the upper bound of UCL that can be achieved by a model because the distributions of word pairs are identical.  

\paragraph{a) Effect of Domains}
To form the English/Fake-English pairs with different distributions, we use the New Testament (\verb|NT|) for English and \{\verb|NT|, Old Testament (\verb|OT|), Wikipedia (\verb|WIKI|), TedTalk (\verb|TED|)\} for Fake-English. Therefore, there are four decipherment conditions with different domains, \verb|NT-NT|, \verb|NT-OT|, \verb|NT-TED| and \verb|NT-WIKI|. \verb|NT| is chosen for its distinct style and therefore word distribution. For both \verb|NT| and \verb|OT| we use the Easy-to-Read Version of the English Bible because of its small vocabulary size. The breakdown of the datasets are shown in Table \ref{tab:datasets}, we also list the unigram JS divergence w.r.t. the \verb|NT|. For all datasets, we use the BPE tokenization scheme of 2K vocabulary size for each language. So \verb|NT-NT| will have 4K vocabulary size in total, 2K for \verb|NT| and 2K for the fake \verb|NT|. 

\begin{table}[ht]
    \centering
    \resizebox{\linewidth}{!}{
    \begin{tabular}{@{}llllll@{}}
    \toprule
    Datasets & Lines & Words & Avg Len & Vocab Size & JS Div.\\ 
    \midrule
    NT & 17K & 202K & 11.9 & 5196 & 0\\
    OT & 17K & 192K & 11.4 & 7644 & 0.0599\\
    TED & 17K & 270K & 15.4 & 12.8K & 0.111\\
    WIKI & 17K & 254K & 14.9 & 15.5K & 0.124\\ 
    \bottomrule
    \end{tabular}
    }
    \caption{\label{tab:datasets} The summary of datasets used.}
\end{table}

\paragraph{b) Effect of Lexical Granularity}
The experiments with different tokenization, which determines the vocabulary size as well as the token granularity, serves two purposes. 

First, they demonstrate the bias introduced by the tokenizer that will directly affect the underlying distribution of the two corpora. As the vocabulary size decreases, its statistical distribution will also be more invariant across different domains, and at the character level will reduce the decipherment problem again to the simplest case of character substitution cipher whereby the decipherment rate with sufficient data is perfect. 

Second, they also simulate the 1-N mapping between a token in the source language with multiple tokens of finer granularity in the target language. Similar to how agglutinative languages will introduce more subwords of finer granularity when constrained to the same vocabulary size. Such behaviour again alters the underlying distribution. We use the language pair \verb|NT-NT| and fixed the 2K vocabulary size for \verb|NT| in the source. For the fake \verb|NT|, we use BPE vocabulary sizes of 500, 1K and 4K. For decipherment conditions of varying lexical granularity we have \verb|NT-500|, \verb|NT-1K| and \verb|NT-4K|. 

\paragraph{c) Effect of Token Order}
Varying token order within a sentence for a given dataset will not change the unigram distribution. However, language modeling techniques generally capture higher-order dependencies, thus, different token order presents a harder challenge for language models. Again we use \verb|NT-NT| and inverse the order of the fake NT to get \verb|NT-INV|. Similarly, we randomize the order of fake NT to get \verb|NT-RND|.

\paragraph{d) Effect of Learning Objectives}

Besides the popular MLM objective used in joint training, we include other possible language modeling objective functions such as causal language modeling (CLM), masked language modeling (MLM) and permutation language modeling (PLM). Although they do not affect the input data distribution, we include them for a more comprehensive analysis. For CLM, we include RNN based implementation, as well as the transformer-based implementation to control the effect of architectural difference. 

\subsection{Model Implementation}
We use the implementations from HuggingFace \cite{wolf-etal-2020-transformers}. 
Both CLM and MLM are based on the BERT \cite{devlin-etal-2019-bert} model.
For PLM we try to keep to the same amount of model parameters. RNN model is adapted from the Fairseq\footnote{\url{https://github.com/pytorch/fairseq}} implementation.
For statistically significant results, we conduct 7 runs with random seeds for each modeling objective under each decipherment condition. In total there are 9 conditions corresponding to 9 datasets, \verb|NT-NT|, \verb|NT-OT|, \verb|NT-TED|, \verb|NT-WIKI|, \verb|NT-500|, \verb|NT-1K|, \verb|NT-4K|, \verb|NT-INV| and \verb|NT-RND|. 

\subsection{Evaluation Metrics}
\label{sec:eval_metrics}
BLI precision is a direct measure of the decipherment performance. However, to evaluate UCL performance, we need to assess the model beyond token mapping. Following the common practices, we include the contextual and sentence-level evaluation. 
With three sets of easily obtainable labels; 1) token to token ground truth correspondence, 2) sentence-level word alignments, 3) sentence retrieval labels, we evaluate the model using BLI, contextual alignment and sentence retrieval to assess the quality of word translation, sequential tagging and sentence classification. Collectively they are used to represent the UCL performance. 

\paragraph{Token translation (BLI)}
is used to evaluate the quality of the cross-lingual embedding transformation, and according to \S\ref{sec:neural_decipher}, it constitutes the language decipherment performance. 
By comparing cosine similarity of the embedding space $\mathbf{E}_f$ and $\mathbf{E}_e$, the $M$ nearest target tokens in $V_f$ for each token in $V_e$ are retrieved.
The retrieved pairs are compared against the ground truth dictionary
which in this case are the English and its prefixed Fake English entries, e.g.\ (cat, ::cat). Performance is measured by precision at $M$. Following a standard evaluation practice \cite{vulic-moens-2013-study, mikolov-etal-2013-exploiting, lample-etal-2018-unsupervised} we report Precision at 1 score (P@1).

\citet{kementchedjhieva-etal-2019-lost} and \citet{glavas-etal-2019-properly} argue the danger of using BLI without in-depth linguistic analysis due to the noise in some dictionaries rendering it not indicative of the actual cross-lingual performance in CLWE. 
\citet{sogaard-etal-2018-limitations} also pointed out that languages are not isomorphic, rendering BLI an inadequate measure of CLWE. However, in our controlled setup, we circumvent those two concerns. Also, BLI is used together with other proxy metrics. Furthermore, we see in \S\ref{sec:analysis} it is well correlated with the UCL performance.

\paragraph{Contextual word alignment} 
aims to determine the alignment of words given two parallel sentences, $S_f$ and $S_e$ at model layer $l$. The score indicates tagging performance and assesses the model's ability to transform words of different `languages' into similar representations at deeper layers. The ground truth alignment between the sentence in English and sentence in Fake-English is the identity matrix.  
We take the word representations of $S_f$ and $S_e$, and in the case of subwords, take the average to represent the whole word \citet{dufter-schutze-2020-identifying}. The alignment is the nearest neighbour of these two representation matrices and compared with the ground truth alignment to get the F1 score averaged across layers $0, 4, 8, 12$.

\paragraph{Sentence retrieval}
is another metric for evaluating cross-lingual representation \cite{artetxe-schwenk-2019-massively}. We obtain the sentence representation by averaging token representations minus start and end tokens in the sentence. One thousand sentence pairs in English and Fake-English datasets, that are not seen by the model during training, are selected. Similar to contextual word alignment, sentences pairs are retrieved based on the nearest cosine distance between the representations of the sentences in target and source. The F1 score is averaged across layers $0, 4, 8, 12$.

\section{Lessons from Decipherment}
\label{sec:analysis}
We find that UCL under joint training paradigm is an interplay of the data domain, input distribution, lexical granularity and modeling objective. And the UCL performance is underpinned by the decipherment difficulty. 
 
\paragraph{Joint learning enables UCL}
\begin{figure}[ht]
    \centering
    \includegraphics[width=\linewidth]{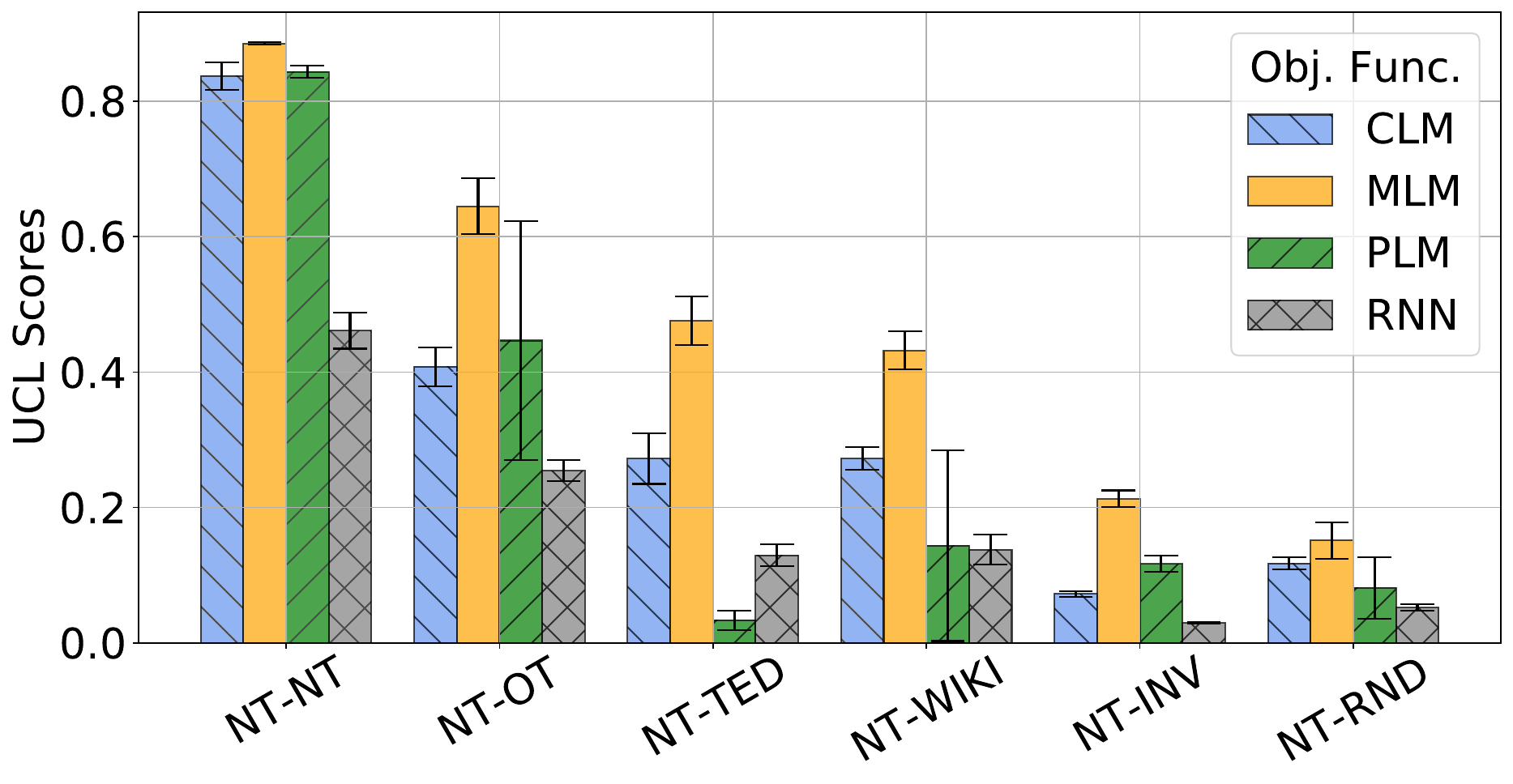}
    \caption{\label{fig:clm_mlm_score} The UCL performance of different objective functions under various decipherment settings.}
\end{figure}
The UCL technique is not unique to mBERT, it is a common characteristic of jointly trained models. However, the effectiveness of UCL varies most notably with the input data distribution between the two corpora (languages). 
All types of objective functions, under the ideal case of \verb|NT-NT| can achieve high UCL scores, Figure~\ref{fig:clm_mlm_score}. Models that are trained separately will certainly have random word-level alignment and near-zero UCL scores.

We can therefore argue that joint training enables the model to use the corpus distribution as a form of cross-lingual signal, as opposed to the usual word, sentence and document level signal.   
The results support that joint learning, which is fundamentally a mutual decipherment process, forms the basis for UCL.

\paragraph{MLM is more tolerant than others}
Different objective functions such as autoregressive, causal, denoising or permutation modeling all achieve notable UCL scores. However, MLM is the most robust as compared to CLM or other sequential types, while CLM deteriorates drastically when a different dataset with divergent distribution is used. Refer to the Figure~\ref{fig:clm_mlm_score}. We compare CLM and MLM that share the same transformer architecture. In CLM the sequential order is enforced through the causal attention mask (\ref{equ:clm}) and in MLM the objective follows a denoising auto-encoder~\cite{vincent-etal-2008-extracting} (\ref{equ:mlm}).
\begin{align}
    \label{equ:clm}
    \text{CLM}(t) = & -\sum_{i} \log p_{\theta} (t_i | t_{< i}) \\
    \label{equ:mlm}
    \text{MLM}(t) = & -\log p_{\theta} (t_m | t_{\neg m}) 
\end{align}
In MLM, $t_m$ are randomly drawn from the sentence to be the target labels and the remaining tokens are the corrupted input $t_{\neg m}$.
Intuitively, MLM induces distributional representation, with less regard to the order information. It is similar in spirit to the bag of word method.
On the other hand, CLM induces sequential distribution representation, akin to the n-gram model. 
Although both representations can be used to align the embedding space between two different domains, CLM is more sensitive to changes in sequence.
In the initial stage of the decipherment task, whereby the model has not recovered any meaningful word correspondence, we need an objective function that can tolerate the misalignments and award any slight improvements so that the model can eventually converge. Otherwise, there can be large swings in the error loss resulting in instability.

\begin{figure}[t]
    \centering
    \includegraphics[width=\linewidth]{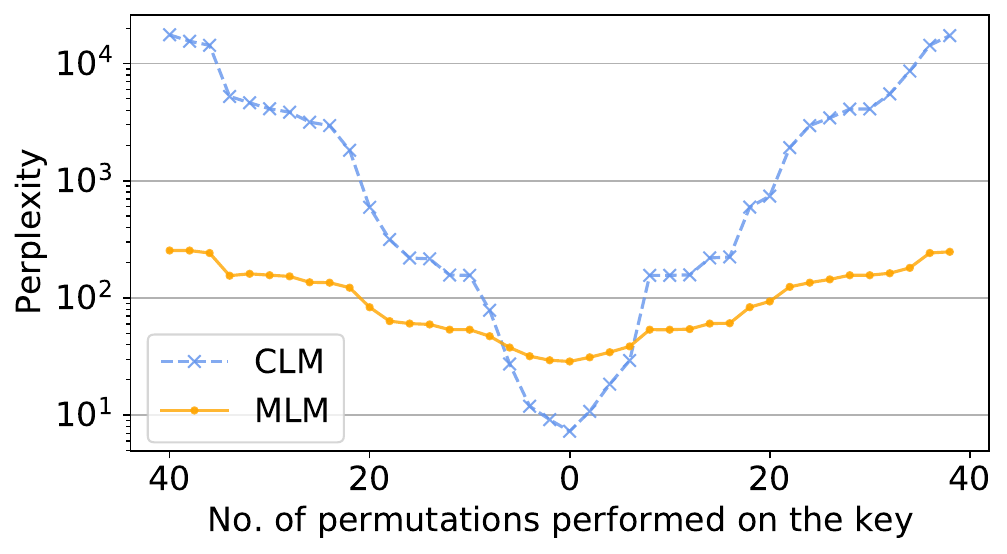}
    \caption{\label{fig:clm_mlm_key}The perplexity of the CLM and MLM model w.r.t. the number of permutations. Zero permutation corresponds to the original validation set, each permutation is a pair-wise swap of two randomly selected vocabulary indices.}
\end{figure}

To quantify this, we used two trained models with the exact setup differing only w.r.t. the objective functions (MLM vs CLM), and evaluate the perplexity on the same validation dataset. To simulate the loss curve w.r.t. the different stages of word decipherment, we take the original vocabulary index and iteratively performs pair-wise permutation. The validation dataset will be indexed using the permutated key. On right and left sides of Figure~\ref{fig:clm_mlm_key}, the model is in the process of deciphering and converging to the ground-truth at the centre. We see that MLM has more tolerant loss curve compared to the sharp rises in CLM that result in swings between the losses.
In practical settings, MLM should be more stable to the sequential discrepancies between the languages. 

\paragraph{UCL score correlates with corpora divergence}
We observe that the divergence in dataset distribution correlates with the UCL score. This is because BLI is directly linked to how closely the contextual distributions agree between the source and target lexicons. As the source and target distributions differ, the BLI performance naturally decrease. Since BLI is a measure of UCL and highly correlated to the overall UCL, the UCL performance is strongly hinged on favourable decipherment conditions, namely statistical similarity. To quantify corpora divergence we use the Jensen-Shannon divergence that considers the unigram distribution.
\begin{figure}[t]
    \centering
    \includegraphics[width=\linewidth]{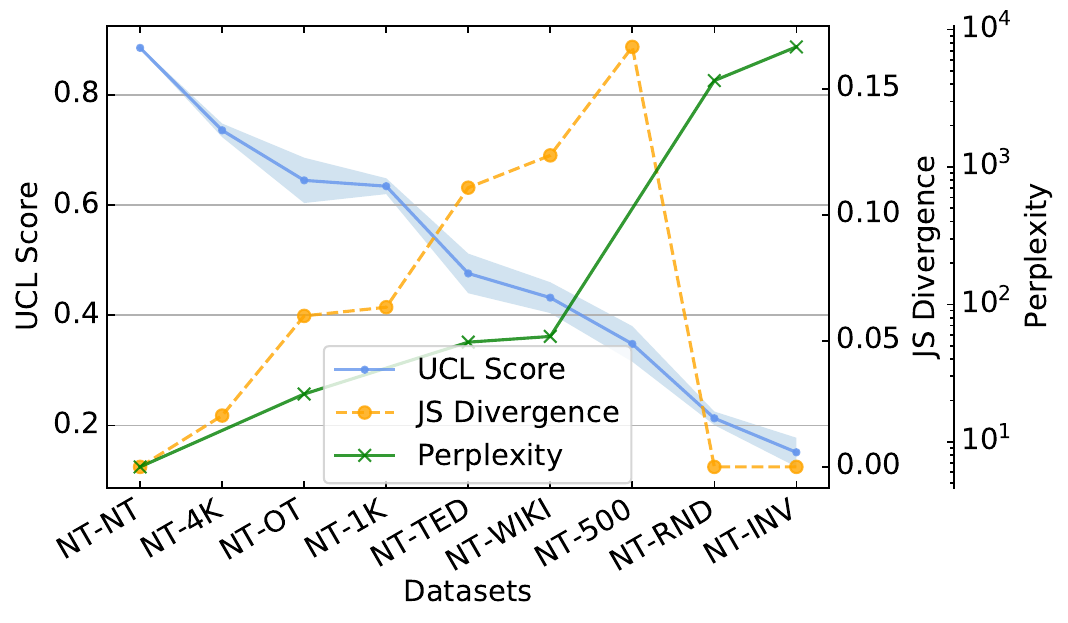}
    \caption{\label{fig:mlm_js} Multilingual score w.r.t. JS divergence based on unigram distribution}
\end{figure}
However, the unigram distribution is not indicative of the actual divergence for datasets with different ordering. For those cases, we use a model trained on the source dataset and perform perplexity evaluation on the target dataset. The perplexity scores consider the n-gram divergence. Our observations are consistent. 

We also want to highlight the effect of tokenization granularity on UCL score. Using compatible granularity on both dataset can establish the one-one correspondence between tokens in $V_e, V_f$. This is the ideal case of \verb|NT-NT|, resulting in the lowest JS score and the best UCL performance. A different granularity on $f$ will introduce more or less subwords in the cases of \verb|NT-4K|, \verb|NT-1K| and \verb|NT-500|. This create one-N or N-N mappings between tokens and makes the decipherment process harder. Therefore, we should establish the appropriate granularity on both languages and strive towards a one-one token correspondence between languages.

\paragraph{The limit of UCL}
\label{sec:limit}
\begin{figure}[ht]
    \centering
    \includegraphics[width=0.8\linewidth]{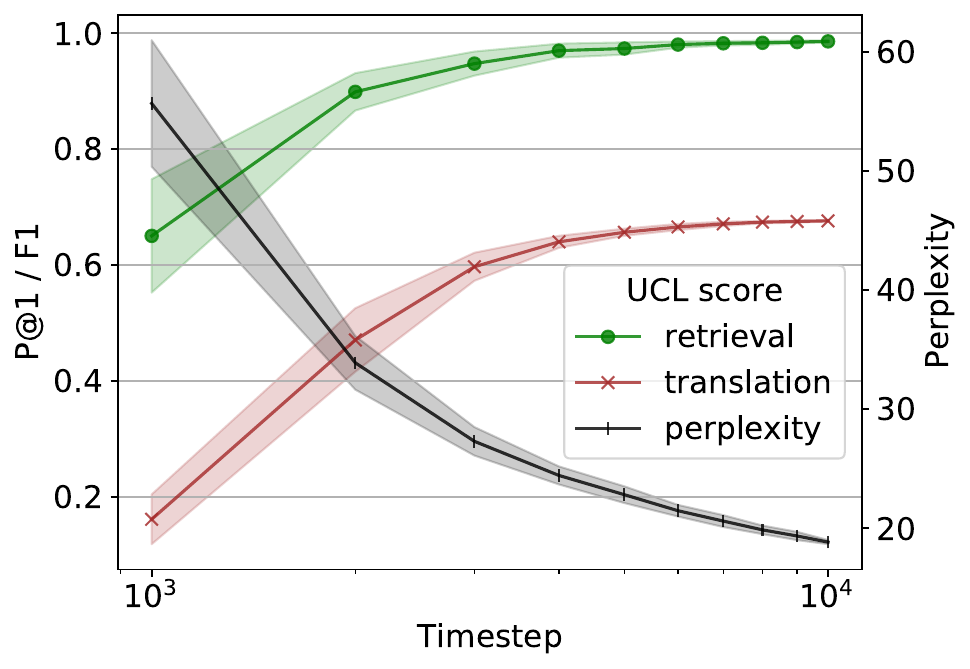}
    \caption{\label{fig:ppl_pos_mlm} The retrieval and BLI performance of a MLM model during training, under NT-NT setting.}
\end{figure}
We observe that perplexity and multilingual scores show an inverse relationship. More specifically, the retrieval and BLI scores improve with the model perplexity as the training progresses, Figure~\ref{fig:ppl_pos_mlm}. This relationship reveals the limit of BLI under unsupervised setup.
\citet{shannon-1949-communication} quantified the inherent uncertainty of the key, $K$, which directly translate to the BLI score in our setup. More importantly, the entropy of the decoded message, i.e. the source language $e$ given the ciphertext, i.e. the target language $f$ is directly linked to the entropy of $K$. Derivation refers to \citet{ravi-knight-2008-attacking}.
\begin{align}
\label{equ:entropy_Kf}
    H(K|f)&  = H(e|f) + C \\
\label{equ:entropy_ef}
    H(e|f)&  = - \sum_e p(e|f) \cdot \log p(e|f)  
\end{align}
We define the entropy $H(e|f)$ in (\ref{equ:entropy_ef}).
Applying to a jointly trained model, $H(e|f)$ is the entropy of the source language model evaluated against the target language and vice versa. $H(e|f)$ is therefore traced by the perplexity curve in Figure~\ref{fig:ppl_pos_mlm}. 

On the other hand, we have the entropy of the key $H(K|f)$, which is $H(e|f)$ added with a positive constant $C$. Since the perplexity can never decrease to zero for any practical system, we have a lower bond for $H(K|f)$ that is always above the perplexity curve. The BLI score is inversely related to $H(K|f)$ and a perfect BLI score is when the entropy of the key is zero. Since $H(K|f)$ is never zero, even in an ideal \verb|NT-NT| setup, thus the BLI score has an upper bound. 

\paragraph{Improvement in decipherment can be transferred to UCL}
The result above has a more profound impact on the overall UCL score because retrieval score and BLI score are highly correlated. To find similar sentences, the encoder relies on the quality of the input token embeddings. The retrieval score is always higher than the BLI score for the same model under the same setting because the selection scope is smaller. 
This also means that we can use a bilingual dictionary to boost both the translation (BLI) as well as the retrieval (Ret) performance. This simple method can overcome the upper limit posed by the $K$ entropy as well as to ease the decipherment difficulty posed by the statistically divergent token pairs. 

To incorporate a dictionary, we introduce an additional MSE loss (\ref{equ:mse}) that minimizes the mean distance between embeddings $(x^e, x^f)$ selected from $\mathbf{E}_e, \mathbf{E}_f$ using $v$ dictionary pairs. This is the \verb|Align| step.
In this English-Fake-English setting, we can ignore the nuances of token alignment across languages, and instead presume the establishment of the one-one token correspondence which is not true between most languages.
In \S\ref{sec:expt_setup}, we will discuss the effect of token correspondence in different languages. 
\begin{equation}
\label{equ:mse}
\Omega_{MSE}  = \frac{1}{v}\sum_{i=1}^{v}(x_i^{e} - {x}_i^{f})^2
\end{equation}

\begin{figure}[ht]
    \centering
    \includegraphics[width=\linewidth]{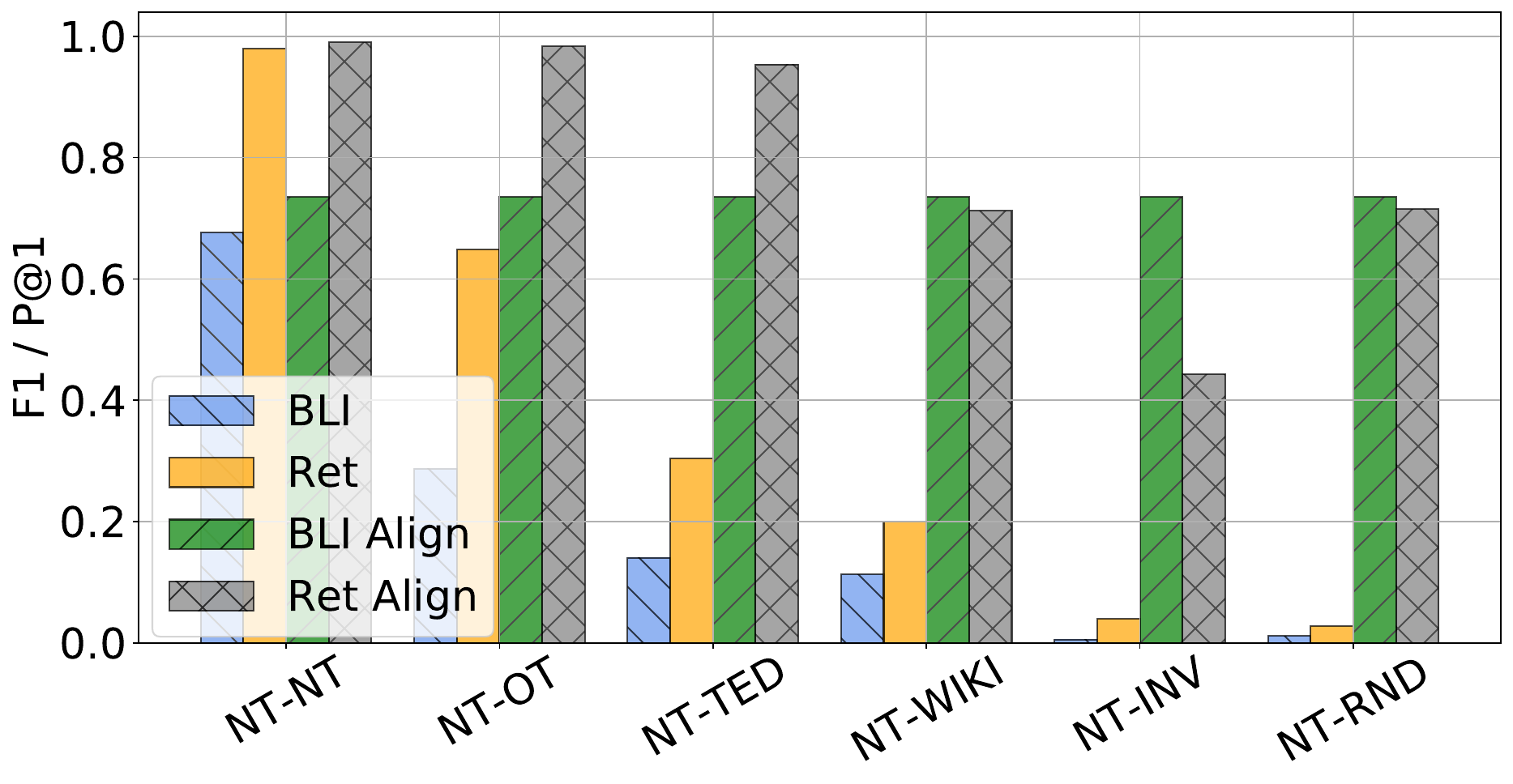}
    \caption{\label{fig:clm_mlm_bli} The BLI performance of different objective functions under various decipherment scenarios.}
\end{figure}
The retrieval performance improves drastically after the \verb|Align| step, matching the performance of an ideal \verb|NT-NT| setting, shown in Figure~\ref{fig:clm_mlm_bli}. This means that for non-ideal decipherment scenarios like the situation between most languages, it is hard to rely on the corpora alone for a robust alignment. 
And the dictionary could be a good bridge between the tokens pairs that have different distributions. 

Similarly in CWLE, supervised methods that make use of dictionaries are usually better than unsupervised methods, especially between distant languages. Even in unsupervised methods, a candidate dictionary is usually explicitly proposed and adapted~\cite{artetxe-etal-2018-robust, artetxe-etal-2017-learning}. We also note the argument that BLI will hurt downstream performance. However, when a dictionary upholds the one-one correspondence between tokens, we don't observe performance degradation. We believe that the selection and construction of token pairs in the dictionary is crucial and should as much as possible uphold the one-one assumption. 
Next, we apply the method to mBERT and show that the cross-lingual performance improves.

\section{Application}
\label{sec:expt_setup}
We take a pretrained m-BERT\footnote{\url{https://huggingface.co/models}} and fine-tune it at the lexical level using a bilingual dictionary. We choose a set of diverse languages, consisting of \verb|de|, \verb|en|, \verb|es|, \verb|fr|, \verb|ja| and \verb|ru|. We take an English-centric pairing due to the ease of interpretation and form the following bi-BERT models, \verb|en-de|, \verb|en-es|, \verb|en-fr|, \verb|en-ja| and \verb|en-ru|. There are possible side-effects of such methodology~\cite{anastasopoulos-neubig-2020-cross}, however, they are orthogonal to the arguments presented in this paper. The high quality dictionaries are extracted from the DBnary project~\cite{serasset-gilles-2015-dbnary} which are based on Wiktionary\footnote{\url{https://www.wiktionary.org/}}. Our modified version\footnote{On average there are 15K entries per dictionary.} is open-sourced. In the fine-tuning stage, we use two objective functions. One optimizes the MSE (\ref{equ:mse}) of the embeddings in the dictionary, another objective continues the language modeling pre-training. 

\paragraph{Control}
To isolate the effect of the MSE objective using a bilingual dictionary, we fine-tune a similar set of bilingual models using only the language modeling objective on the same corpora. In \S\ref{sec:discussion}, we only report the difference in performance ($\Delta$ UCL) compared to these control models.

\subsection{Experiments}
We note that~\cite{conneau-etal-2020-emerging} conducted similar experiments to study the most important factor in cross-lingual learning. We conducted a more comprehensive setup with the targeted embedding alignment. In our experiment, we subdivide the dictionary into different syntactic groups and investigate how each group will affect cross-lingual learning differently.
We limit our experiment to a bilingual setting because multilingual dictionaries have the additional problem of triangulation and incompatibility issues. The downstream tasks are a subset of XTREME \cite{hu-etal-2020-xtreme} benchmarking that is designed to evaluate the model's zero-shot cross-lingual performance. We select the classification, structured prediction and retrieval tasks that are analogous to translation, contextual alignment and retrieval metrics proposed in \S~\ref{sec:eval_metrics}.

\paragraph{Classification}
The classification task is made up of XNLI and PAWS-X.
The XNLI task requires the model to decide whether the premise sentence entails, contradicts, or is neutral toward the hypothesis sentence. For training, the MultiNLI \cite{williams-etal-2018-broad} training data is used and the dataset contains only English examples. The trained model will be evaluated on other languages using the Cross-lingual Natural Language Inference corpus \cite{conneau-etal-2018-xnli} 
(\verb|ja| is not included in the dataset).
Similar to XNLI, the Cross-lingual Paraphrase Adversaries from Word Scrambling \cite{yang-etal-2019-paws} (PAWS-X) dataset asks the model to make a binary decision whether the two sentences are paraphrases. Training is done using only the English examples from PAWS \cite{zhang-etal-2019-paws} and the evaluation for other languages (\verb|ru| is not included in the dataset) are done using the translation of the dev and test subsets. 
Both PAWS-X and XNLI are useful for evaluating the quality of the cross-lingual sentence representation.

\paragraph{Structured Prediction}
The POS tagging dataset is from the Universal Dependencies v2.5 \cite{nivre-etal-2020-universal} treebanks which consists of 17 POS tags. The training data is in English, the model is evaluated on the target languages.
The NER dataset is based on Wikiann \cite{pan-etal-2017-cross}
in which named entities in Wikipedia were automatically annotated with LOC, PER, and ORG tags in IOB2 format. In the zero-shot transfer evaluation a balanced train, dev, and test splits \cite{rahimi-etal-2019-massively} is used. All target languages are present for both POS and NER tasks.

\paragraph{Retrieval}
The retrieval task consists of \textasciitilde 1,000 English-aligned sentence pairs from the Tatoeba dataset \cite{artetxe-schwenk-2019-massively}. The nearest neighbour is retrieved based on cosine similarity in the target language and the error rate is reported. All target languages are present for this task.

\subsection{Discussion}
\label{sec:discussion}
\paragraph{Alignment improves performance}
\begin{figure}[ht]
    \centering
    \includegraphics[width=\linewidth]{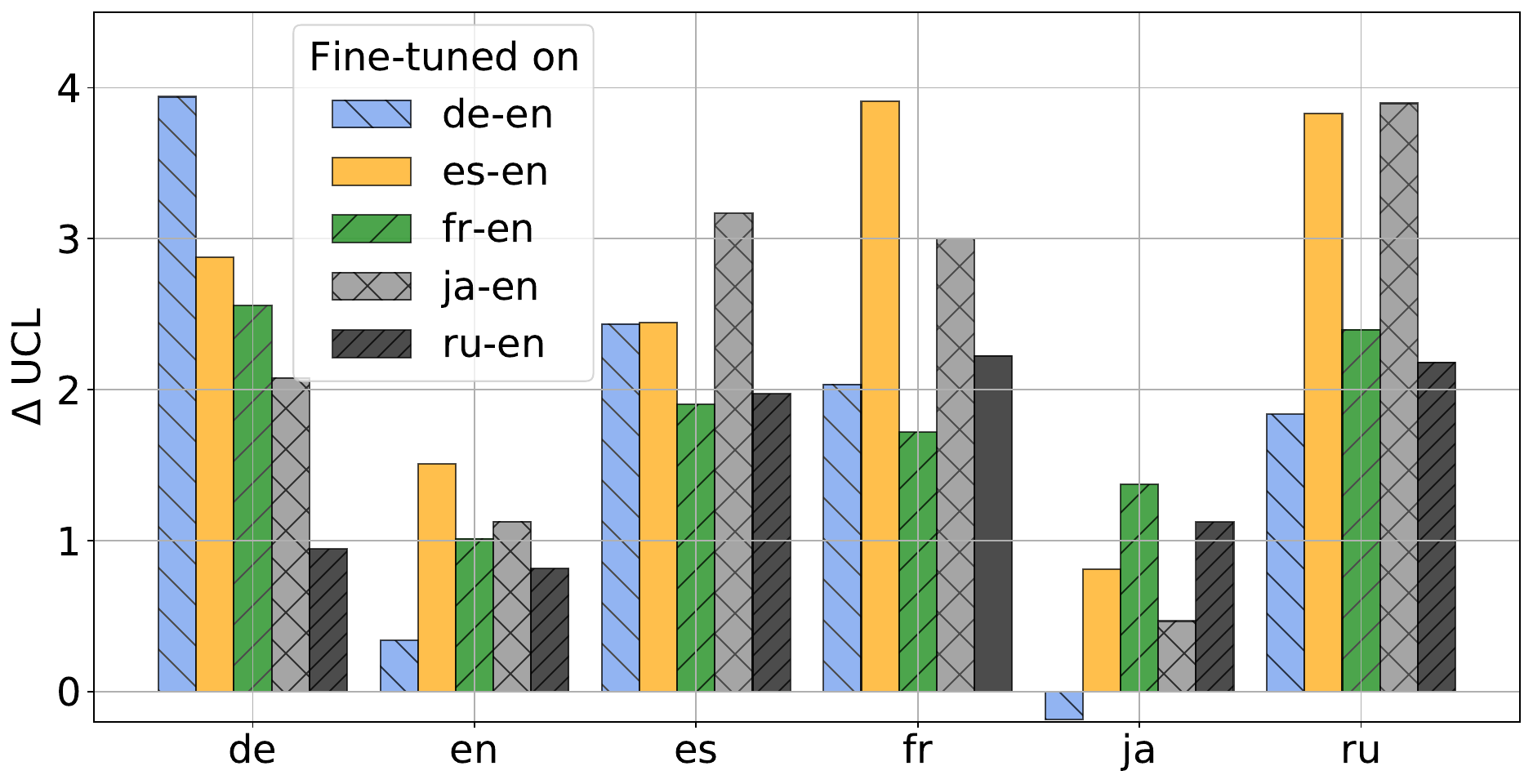}
    \caption{\label{fig:ave_by_lan} The difference in zero-shot cross-lingual transfer performance compared to the control model. Each model, which is fine-tuned on the bilingual dictionary, f-e, is evaluated on five other languages.}
\end{figure}

We show that UCL is strongly tied to the decipherment task, which a jointly trained model is implicitly optimizing. In the controlled setup, we showed that improvements in decipherment transfer to UCL, this also suggest that UCL suffers from divergent statistics and a lack of one-one token correspondence. Although a bilingual dictionary does not strictly obey the one-one mapping, it still helps in the decipherment process to bridge the language distance. In Figure~\ref{fig:ave_by_lan}, taking the model fine-tuned on \verb|de-en| for instance, we observe a significant increment of $3.94$ in absolute score. Although only the \verb|de-en| dictionary is used, not only does the transfer learning in \verb|en| $\rightarrow$ \verb|de| improves, other languages which are not fine-tuned improve as well. \verb|fr|, \verb|ru| and \verb|es| benefit the most which could indicate an inherent relationship between those languages. Similarly for \verb|ru-en|, apart from the most significant improvement in \verb|ru|, \verb|fr| and \verb|de| also improved. The average performance improves by $2.29$ in absolute score. It is significant for the language of interest, and the improvement transfers to related languages without significantly compromising the performance on other languages. 

\paragraph{Different responses to word types}
To examine how well those different categories of words align across languages, we divide the dictionary according to its POS tags into 4 groups. Nouns, proper nouns and numerals are grouped into \verb|Noun|, verbs into \verb|Verb| and adjectives and adverbs into \verb|Adj/Adv|. Lastly, functional words such as prepositions, conjunctions, pronouns and determiners are grouped into \verb|Functional|. Additionally, we include \verb|All| which consists of all the entries in the dictionary and \verb|Whole word| which comprises \verb|All| minus the subwords, for comparison. 
\begin{figure}[ht]
    \centering
    \includegraphics[width=\linewidth]{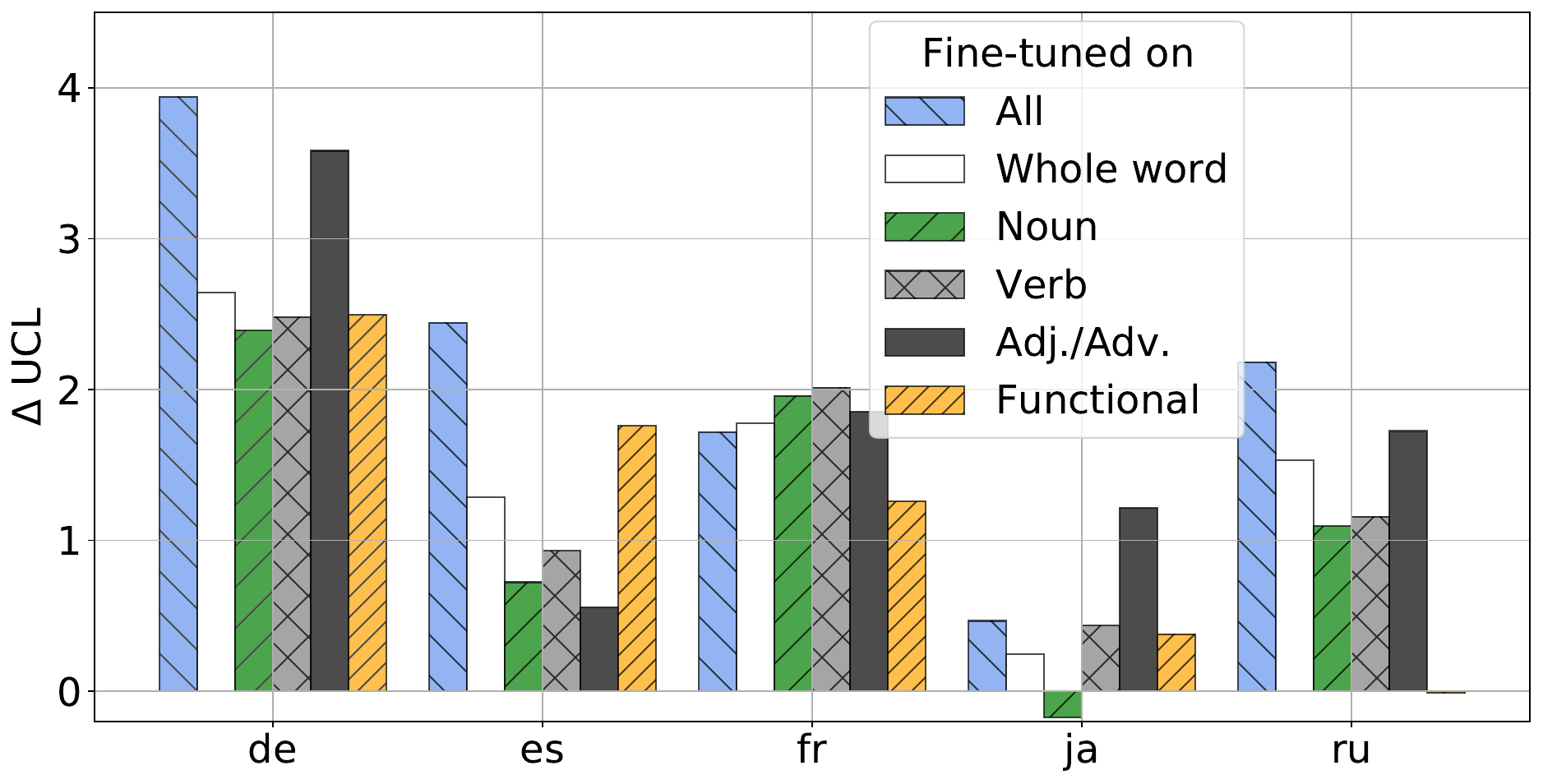}
    \caption{\label{fig:dict_by_lan} The difference in zero-shot cross-lingual transfer performance compared to the control model. For each language we the performance difference w.r.t.\ the aligned word type is shown.}
\end{figure}
The results in Figure~\ref{fig:dict_by_lan} show that choosing different word types to align will affect the cross-lingual performance. The bars within the same language cluster indicate the target language's performance responses for different subsets of dictionaries. We observe that while aligning all the words in the dictionary gives the most performance boost, aligning just the adjectives and adverbs within the dictionary gives competitive results in all languages except for \verb|es| and \verb|ja|. The reason could be that adjective and adverbs are more consistent in their inflexions, thus the subwords formed due to tokenization are easily aligned between English and the target languages. Another reason could be the consistent usage of adjectives and adverbs across English and the target language. 
While there are common characteristics like the similar performance between aligning \verb|Noun| and \verb|Verb|, other observations vary between languages, indicating the need to analyse word behaviours specific for that language pair. 

Aligning \verb|Noun| group for \verb|ja| deteriorates the performance possibly due to the breakdown of Japanese words into single hiragana or katakana syllabaries during the tokenization process. Across the five languages, \verb|ja| has the lowest number of \verb|Whole word| of $753$ entries which is \textasciitilde 4 times less than other dictionaries. The greater number of subwords in Japanese and the different ways of subword formation indicate incompatible tokenization granularity and harder alignment. We breakdown the \verb|de-en| dictionary by word to subword ratio (and number of entries); \verb|Noun|: $0.115$ (7.6K), \verb|Verb|: $0.0787$ (3.2K), \verb|Adj/Adv|: $0.275$ (2.0K), \verb|Functional|: $2.46$ (0.3K). The better performance of \verb|Adj/Adv| may be due to its higher granularity agreement (more whole words) and the establishment of more one-one word pairs. This observation highlights one of the major difficulties in aligning the two languages; whether each token in the source language can be attributed to a counterpart in the target language.  

\paragraph{Word type contributions to task categories}
\begin{figure}[ht]
    \centering
    \includegraphics[width=\linewidth]{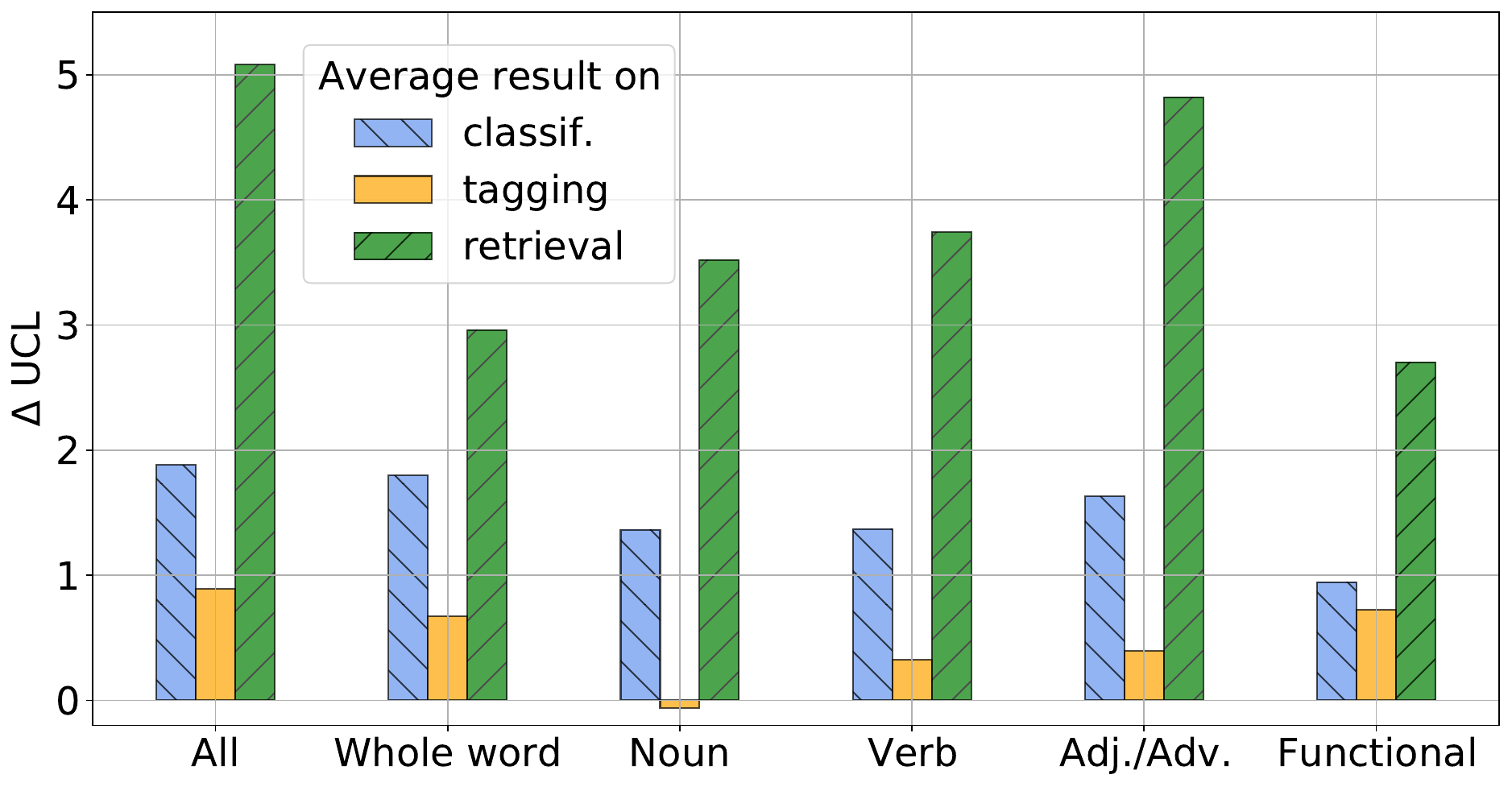}
    \caption{\label{fig:dict_by_task} The difference in zero-shot cross-lingual transfer performance compared to the control model. For aligning each types of word we plot w.r.t.\ the task categories.}
\end{figure}
Lexical alignment as a way of easing the decipherment difficulty significant improves the retrieval task in which the \verb|Adj/Adv| group still fares well. As the word alignment improves, the sentential representations also tend to be aligned better. The improvement in classification is notably lower across the groups and the reason could be the need to rely on higher-level signals than just lexical correspondence to make a decision. For structured prediction, we see that aligning functional words is as good as aligning the whole dictionary, which is again expected because the functional words strongly suggest the syntax of words. Another explanation for the lower improvement is due to the dictionary being pair-wise entries without additional syntax information to further disambiguate different senses. 

\section{Conclusion}
This paper argues for a decipherment-aware multilingual learning perspective. The overarching theme of the paper is to establish the role of decipherment in multilingual learning. We argue the importance of lexical correspondence between language pairs and how tokenization scheme and corpora distribution affect the one-one assumption and ultimately the multilingual learning performance. 
We also urge for a more language-pair-specific analysis to bridge their topological difference. A possible future direction is to incorporate explicit mechanism to tackle the many-many token alignments.

% Bibliography entries for the entire Anthology, followed by custom entries
%\bibliography{anthology,custom}
% Custom bibliography entries only
\bibliography{anthology, custom}

\begin{thebibliography}{59}
\providecommand{\natexlab}[1]{#1}

\bibitem[{Anastasopoulos and Neubig(2020)}]{anastasopoulos-neubig-2020-cross}
Antonios Anastasopoulos and Graham Neubig. 2020.
\newblock \href {https://doi.org/10.18653/v1/2020.acl-main.766} {Should all
  cross-lingual embeddings speak {E}nglish?}
\newblock In \emph{Proceedings of the 58th Annual Meeting of the Association
  for Computational Linguistics}, pages 8658--8679, Online. Association for
  Computational Linguistics.

\bibitem[{Artetxe et~al.(2017)Artetxe, Labaka, and
  Agirre}]{artetxe-etal-2017-learning}
Mikel Artetxe, Gorka Labaka, and Eneko Agirre. 2017.
\newblock \href {https://doi.org/10.18653/v1/P17-1042} {Learning bilingual word
  embeddings with (almost) no bilingual data}.
\newblock In \emph{Proceedings of the 55th Annual Meeting of the Association
  for Computational Linguistics (Volume 1: Long Papers)}, pages 451--462,
  Vancouver, Canada. Association for Computational Linguistics.

\bibitem[{Artetxe et~al.(2018{\natexlab{a}})Artetxe, Labaka, and
  Agirre}]{artetxe-etal-2018-robust}
Mikel Artetxe, Gorka Labaka, and Eneko Agirre. 2018{\natexlab{a}}.
\newblock \href {https://doi.org/10.18653/v1/P18-1073} {A robust self-learning
  method for fully unsupervised cross-lingual mappings of word embeddings}.
\newblock In \emph{Proceedings of the 56th Annual Meeting of the Association
  for Computational Linguistics (Volume 1: Long Papers)}, pages 789--798,
  Melbourne, Australia. Association for Computational Linguistics.

\bibitem[{Artetxe et~al.(2018{\natexlab{b}})Artetxe, Labaka, and
  Agirre}]{artetxe-etal-2018-unsupervised}
Mikel Artetxe, Gorka Labaka, and Eneko Agirre. 2018{\natexlab{b}}.
\newblock \href {https://doi.org/10.18653/v1/D18-1399} {Unsupervised
  statistical machine translation}.
\newblock In \emph{Proceedings of the 2018 Conference on Empirical Methods in
  Natural Language Processing}, pages 3632--3642, Brussels, Belgium.
  Association for Computational Linguistics.

\bibitem[{Artetxe et~al.(2020{\natexlab{a}})Artetxe, Ruder, and
  Yogatama}]{artetxe-etal-2020-cross}
Mikel Artetxe, Sebastian Ruder, and Dani Yogatama. 2020{\natexlab{a}}.
\newblock \href {https://doi.org/10.18653/v1/2020.acl-main.421} {On the
  cross-lingual transferability of monolingual representations}.
\newblock In \emph{Proceedings of the 58th Annual Meeting of the Association
  for Computational Linguistics}, pages 4623--4637, Online. Association for
  Computational Linguistics.

\bibitem[{Artetxe et~al.(2020{\natexlab{b}})Artetxe, Ruder, Yogatama, Labaka,
  and Agirre}]{artetxe-etal-2020-call}
Mikel Artetxe, Sebastian Ruder, Dani Yogatama, Gorka Labaka, and Eneko Agirre.
  2020{\natexlab{b}}.
\newblock \href {https://doi.org/10.18653/v1/2020.acl-main.658} {A call for
  more rigor in unsupervised cross-lingual learning}.
\newblock In \emph{Proceedings of the 58th Annual Meeting of the Association
  for Computational Linguistics}, pages 7375--7388, Online. Association for
  Computational Linguistics.

\bibitem[{Artetxe and Schwenk(2019)}]{artetxe-schwenk-2019-massively}
Mikel Artetxe and Holger Schwenk. 2019.
\newblock \href {https://doi.org/10.1162/tacl_a_00288} {Massively multilingual
  sentence embeddings for zero-shot cross-lingual transfer and beyond}.
\newblock \emph{Transactions of the Association for Computational Linguistics},
  7:597--610.

\bibitem[{Conneau et~al.(2020{\natexlab{a}})Conneau, Khandelwal, Goyal,
  Chaudhary, Wenzek, Guzm{\'a}n, Grave, Ott, Zettlemoyer, and
  Stoyanov}]{conneau-etal-2020-unsupervised}
Alexis Conneau, Kartikay Khandelwal, Naman Goyal, Vishrav Chaudhary, Guillaume
  Wenzek, Francisco Guzm{\'a}n, Edouard Grave, Myle Ott, Luke Zettlemoyer, and
  Veselin Stoyanov. 2020{\natexlab{a}}.
\newblock \href {https://doi.org/10.18653/v1/2020.acl-main.747} {Unsupervised
  cross-lingual representation learning at scale}.
\newblock In \emph{Proceedings of the 58th Annual Meeting of the Association
  for Computational Linguistics}, pages 8440--8451, Online. Association for
  Computational Linguistics.

\bibitem[{Conneau and Lample(2019)}]{conneau-lample-2019-cross}
Alexis Conneau and Guillaume Lample. 2019.
\newblock \href
  {https://proceedings.neurips.cc/paper/2019/file/c04c19c2c2474dbf5f7ac4372c5b9af1-Paper.pdf}
  {Cross-lingual language model pretraining}.
\newblock In \emph{Advances in Neural Information Processing Systems},
  volume~32. Curran Associates, Inc.

\bibitem[{Conneau et~al.(2018)Conneau, Rinott, Lample, Williams, Bowman,
  Schwenk, and Stoyanov}]{conneau-etal-2018-xnli}
Alexis Conneau, Ruty Rinott, Guillaume Lample, Adina Williams, Samuel Bowman,
  Holger Schwenk, and Veselin Stoyanov. 2018.
\newblock \href {https://doi.org/10.18653/v1/D18-1269} {{XNLI}: Evaluating
  cross-lingual sentence representations}.
\newblock In \emph{Proceedings of the 2018 Conference on Empirical Methods in
  Natural Language Processing}, pages 2475--2485, Brussels, Belgium.
  Association for Computational Linguistics.

\bibitem[{Conneau et~al.(2020{\natexlab{b}})Conneau, Wu, Li, Zettlemoyer, and
  Stoyanov}]{conneau-etal-2020-emerging}
Alexis Conneau, Shijie Wu, Haoran Li, Luke Zettlemoyer, and Veselin Stoyanov.
  2020{\natexlab{b}}.
\newblock \href {https://doi.org/10.18653/v1/2020.acl-main.536} {Emerging
  cross-lingual structure in pretrained language models}.
\newblock In \emph{Proceedings of the 58th Annual Meeting of the Association
  for Computational Linguistics}, pages 6022--6034, Online. Association for
  Computational Linguistics.

\bibitem[{Devlin et~al.(2019)Devlin, Chang, Lee, and
  Toutanova}]{devlin-etal-2019-bert}
Jacob Devlin, Ming-Wei Chang, Kenton Lee, and Kristina Toutanova. 2019.
\newblock \href {https://doi.org/10.18653/v1/N19-1423} {{BERT}: Pre-training of
  deep bidirectional transformers for language understanding}.
\newblock In \emph{Proceedings of the 2019 Conference of the North {A}merican
  Chapter of the Association for Computational Linguistics: Human Language
  Technologies, Volume 1 (Long and Short Papers)}, pages 4171--4186,
  Minneapolis, Minnesota. Association for Computational Linguistics.

\bibitem[{Dou and Knight(2013)}]{dou-knight-2013-dependency}
Qing Dou and Kevin Knight. 2013.
\newblock \href {https://www.aclweb.org/anthology/D13-1173} {Dependency-based
  decipherment for resource-limited machine translation}.
\newblock In \emph{Proceedings of the 2013 Conference on Empirical Methods in
  Natural Language Processing}, pages 1668--1676, Seattle, Washington, USA.
  Association for Computational Linguistics.

\bibitem[{Dufter and Sch{\"u}tze(2020)}]{dufter-schutze-2020-identifying}
Philipp Dufter and Hinrich Sch{\"u}tze. 2020.
\newblock \href {https://doi.org/10.18653/v1/2020.emnlp-main.358} {Identifying
  elements essential for {BERT}{'}s multilinguality}.
\newblock In \emph{Proceedings of the 2020 Conference on Empirical Methods in
  Natural Language Processing (EMNLP)}, pages 4423--4437, Online. Association
  for Computational Linguistics.

\bibitem[{Faruqui and Dyer(2014)}]{faruqui-dyer-2014-improving}
Manaal Faruqui and Chris Dyer. 2014.
\newblock \href {https://doi.org/10.3115/v1/E14-1049} {Improving vector space
  word representations using multilingual correlation}.
\newblock In \emph{Proceedings of the 14th Conference of the {E}uropean Chapter
  of the Association for Computational Linguistics}, pages 462--471,
  Gothenburg, Sweden. Association for Computational Linguistics.

\bibitem[{Fung(1995)}]{fung-1995-compiling}
Pascale Fung. 1995.
\newblock \href {https://www.aclweb.org/anthology/W95-0114} {Compiling
  bilingual lexicon entries from a non-parallel {E}nglish-{C}hinese corpus}.
\newblock In \emph{Third Workshop on Very Large Corpora}.

\bibitem[{Glava{\v{s}} et~al.(2019)Glava{\v{s}}, Litschko, Ruder, and
  Vuli{\'c}}]{glavas-etal-2019-properly}
Goran Glava{\v{s}}, Robert Litschko, Sebastian Ruder, and Ivan Vuli{\'c}. 2019.
\newblock \href {https://doi.org/10.18653/v1/P19-1070} {How to (properly)
  evaluate cross-lingual word embeddings: On strong baselines, comparative
  analyses, and some misconceptions}.
\newblock In \emph{Proceedings of the 57th Annual Meeting of the Association
  for Computational Linguistics}, pages 710--721, Florence, Italy. Association
  for Computational Linguistics.

\bibitem[{Haghighi et~al.(2008)Haghighi, Liang, Berg-Kirkpatrick, and
  Klein}]{haghighi-etal-2008-learning}
Aria Haghighi, Percy Liang, Taylor Berg-Kirkpatrick, and Dan Klein. 2008.
\newblock \href {https://www.aclweb.org/anthology/P08-1088} {Learning bilingual
  lexicons from monolingual corpora}.
\newblock In \emph{Proceedings of ACL-08: HLT}, pages 771--779, Columbus, Ohio.
  Association for Computational Linguistics.

\bibitem[{Hu et~al.(2020)Hu, Ruder, Siddhant, Neubig, Firat, and
  Johnson}]{hu-etal-2020-xtreme}
Junjie Hu, Sebastian Ruder, Aditya Siddhant, Graham Neubig, Orhan Firat, and
  Melvin Johnson. 2020.
\newblock \href {https://arxiv.org/abs/2003.11080} {Xtreme: A massively
  multilingual multi-task benchmark for evaluating cross-lingual
  generalization}.
\newblock \emph{CoRR}, abs/2003.11080.

\bibitem[{Huang et~al.(2015)Huang, Gardner, Papalexakis, Faloutsos,
  Sidiropoulos, Mitchell, Talukdar, and Fu}]{huang-etal-2015-translation}
Kejun Huang, Matt Gardner, Evangelos Papalexakis, Christos Faloutsos, Nikos
  Sidiropoulos, Tom Mitchell, Partha~P. Talukdar, and Xiao Fu. 2015.
\newblock \href {https://doi.org/10.18653/v1/D15-1127} {Translation invariant
  word embeddings}.
\newblock In \emph{Proceedings of the 2015 Conference on Empirical Methods in
  Natural Language Processing}, pages 1084--1088, Lisbon, Portugal. Association
  for Computational Linguistics.

\bibitem[{K et~al.(2020)K, Wang, Mayhew, and Roth}]{K-2020-cross-Lingual}
Karthikeyan K, Zihan Wang, Stephen Mayhew, and Dan Roth. 2020.
\newblock \href {https://openreview.net/forum?id=HJeT3yrtDr} {Cross-lingual
  ability of multilingual bert: An empirical study}.
\newblock In \emph{International Conference on Learning Representations}.

\bibitem[{Kementchedjhieva et~al.(2019)Kementchedjhieva, Hartmann, and
  S{\o}gaard}]{kementchedjhieva-etal-2019-lost}
Yova Kementchedjhieva, Mareike Hartmann, and Anders S{\o}gaard. 2019.
\newblock \href {https://doi.org/10.18653/v1/D19-1328} {Lost in evaluation:
  Misleading benchmarks for bilingual dictionary induction}.
\newblock In \emph{Proceedings of the 2019 Conference on Empirical Methods in
  Natural Language Processing and the 9th International Joint Conference on
  Natural Language Processing (EMNLP-IJCNLP)}, pages 3336--3341, Hong Kong,
  China. Association for Computational Linguistics.

\bibitem[{Koehn and Knight(2002)}]{koehn-knight-2002-learning}
Philipp Koehn and Kevin Knight. 2002.
\newblock \href {https://doi.org/10.3115/1118627.1118629} {Learning a
  translation lexicon from monolingual corpora}.
\newblock In \emph{Proceedings of the {ACL}-02 Workshop on Unsupervised Lexical
  Acquisition}, pages 9--16, Philadelphia, Pennsylvania, USA. Association for
  Computational Linguistics.

\bibitem[{Lample et~al.(2018{\natexlab{a}})Lample, Conneau, Denoyer, and
  Ranzato}]{lample-etal-2018-unsupervised}
Guillaume Lample, Alexis Conneau, Ludovic Denoyer, and Marc'Aurelio Ranzato.
  2018{\natexlab{a}}.
\newblock \href {https://openreview.net/forum?id=rkYTTf-AZ} {Unsupervised
  machine translation using monolingual corpora only}.
\newblock In \emph{International Conference on Learning Representations}.

\bibitem[{Lample et~al.(2018{\natexlab{b}})Lample, Conneau, Ranzato, Denoyer,
  and Jégou}]{lample-etal-2018-word}
Guillaume Lample, Alexis Conneau, Marc'Aurelio Ranzato, Ludovic Denoyer, and
  Hervé Jégou. 2018{\natexlab{b}}.
\newblock \href {https://openreview.net/forum?id=H196sainb} {Word translation
  without parallel data}.
\newblock In \emph{International Conference on Learning Representations}.

\bibitem[{Lee and Li(2019)}]{8683678}
Grandee Lee and Haizhou Li. 2019.
\newblock \href {https://doi.org/10.1109/ICASSP.2019.8683678} {Word and class
  common space embedding for code-switch language modelling}.
\newblock In \emph{ICASSP 2019 - 2019 IEEE International Conference on
  Acoustics, Speech and Signal Processing (ICASSP)}, pages 6086--6090.

\bibitem[{Lee and Li(2020)}]{lee-li-2020-modeling}
Grandee Lee and Haizhou Li. 2020.
\newblock \href {https://doi.org/10.18653/v1/2020.acl-main.80} {Modeling
  code-switch languages using bilingual parallel corpus}.
\newblock In \emph{Proceedings of the 58th Annual Meeting of the Association
  for Computational Linguistics}, pages 860--870, Online. Association for
  Computational Linguistics.

\bibitem[{Lee et~al.(2019)Lee, Yue, and Li}]{lee19d_interspeech}
Grandee Lee, Xianghu Yue, and Haizhou Li. 2019.
\newblock \href {https://doi.org/10.21437/Interspeech.2019-1382}
  {{Linguistically Motivated Parallel Data Augmentation for Code-Switch
  Language Modeling}}.
\newblock In \emph{Proc. Interspeech 2019}, pages 3730--3734.

\bibitem[{Levy et~al.(2017)Levy, S{\o}gaard, and
  Goldberg}]{levy-etal-2017-strong}
Omer Levy, Anders S{\o}gaard, and Yoav Goldberg. 2017.
\newblock \href {https://www.aclweb.org/anthology/E17-1072} {A strong baseline
  for learning cross-lingual word embeddings from sentence alignments}.
\newblock In \emph{Proceedings of the 15th Conference of the {E}uropean Chapter
  of the Association for Computational Linguistics: Volume 1, Long Papers},
  pages 765--774, Valencia, Spain. Association for Computational Linguistics.

\bibitem[{Luo et~al.(2019)Luo, Cao, and Barzilay}]{luo-etal-2019-neural}
Jiaming Luo, Yuan Cao, and Regina Barzilay. 2019.
\newblock \href {https://doi.org/10.18653/v1/P19-1303} {Neural decipherment via
  minimum-cost flow: From {U}garitic to {L}inear {B}}.
\newblock In \emph{Proceedings of the 57th Annual Meeting of the Association
  for Computational Linguistics}, pages 3146--3155, Florence, Italy.
  Association for Computational Linguistics.

\bibitem[{Luo et~al.(2021)Luo, Hartmann, Santus, Barzilay, and
  Cao}]{luo-etal-2021-deciphering}
Jiaming Luo, Frederik Hartmann, Enrico Santus, Regina Barzilay, and Yuan Cao.
  2021.
\newblock \href {https://doi.org/10.1162/tacl_a_00354} {Deciphering
  undersegmented ancient scripts using phonetic prior}.
\newblock \emph{Transactions of the Association for Computational Linguistics},
  9:69--81.

\bibitem[{Miceli~Barone(2016)}]{miceli-barone-2016-towards}
Antonio~Valerio Miceli~Barone. 2016.
\newblock \href {https://doi.org/10.18653/v1/W16-1614} {Towards cross-lingual
  distributed representations without parallel text trained with adversarial
  autoencoders}.
\newblock In \emph{Proceedings of the 1st Workshop on Representation Learning
  for {NLP}}, pages 121--126, Berlin, Germany. Association for Computational
  Linguistics.

\bibitem[{Mikolov et~al.(2013)Mikolov, Le, and
  Sutskever}]{mikolov-etal-2013-exploiting}
Tomas Mikolov, Quoc~V Le, and Ilya Sutskever. 2013.
\newblock Exploiting similarities among languages for machine translation.
\newblock \emph{arXiv preprint arXiv:1309.4168}.

\bibitem[{Nivre et~al.(2020)Nivre, de~Marneffe, Ginter, Haji{\v{c}}, Manning,
  Pyysalo, Schuster, Tyers, and Zeman}]{nivre-etal-2020-universal}
Joakim Nivre, Marie-Catherine de~Marneffe, Filip Ginter, Jan Haji{\v{c}},
  Christopher~D. Manning, Sampo Pyysalo, Sebastian Schuster, Francis Tyers, and
  Daniel Zeman. 2020.
\newblock \href {https://www.aclweb.org/anthology/2020.lrec-1.497} {{U}niversal
  {D}ependencies v2: An evergrowing multilingual treebank collection}.
\newblock In \emph{Proceedings of the 12th Language Resources and Evaluation
  Conference}, pages 4034--4043, Marseille, France. European Language Resources
  Association.

\bibitem[{Nuhn et~al.(2012)Nuhn, Mauser, and Ney}]{nuhn-etal-2012-deciphering}
Malte Nuhn, Arne Mauser, and Hermann Ney. 2012.
\newblock \href {https://www.aclweb.org/anthology/P12-1017} {Deciphering
  foreign language by combining language models and context vectors}.
\newblock In \emph{Proceedings of the 50th Annual Meeting of the Association
  for Computational Linguistics (Volume 1: Long Papers)}, pages 156--164, Jeju
  Island, Korea. Association for Computational Linguistics.

\bibitem[{Nuhn et~al.(2013)Nuhn, Schamper, and Ney}]{nuhn-etal-2013-beam}
Malte Nuhn, Julian Schamper, and Hermann Ney. 2013.
\newblock \href {https://www.aclweb.org/anthology/P13-1154} {Beam search for
  solving substitution ciphers}.
\newblock In \emph{Proceedings of the 51st Annual Meeting of the Association
  for Computational Linguistics (Volume 1: Long Papers)}, pages 1568--1576,
  Sofia, Bulgaria. Association for Computational Linguistics.

\bibitem[{Ormazabal et~al.(2019)Ormazabal, Artetxe, Labaka, Soroa, and
  Agirre}]{ormazabal-etal-2019-analyzing}
Aitor Ormazabal, Mikel Artetxe, Gorka Labaka, Aitor Soroa, and Eneko Agirre.
  2019.
\newblock \href {https://doi.org/10.18653/v1/P19-1492} {Analyzing the
  limitations of cross-lingual word embedding mappings}.
\newblock In \emph{Proceedings of the 57th Annual Meeting of the Association
  for Computational Linguistics}, pages 4990--4995, Florence, Italy.
  Association for Computational Linguistics.

\bibitem[{Pan et~al.(2017)Pan, Zhang, May, Nothman, Knight, and
  Ji}]{pan-etal-2017-cross}
Xiaoman Pan, Boliang Zhang, Jonathan May, Joel Nothman, Kevin Knight, and Heng
  Ji. 2017.
\newblock \href {https://doi.org/10.18653/v1/P17-1178} {Cross-lingual name
  tagging and linking for 282 languages}.
\newblock In \emph{Proceedings of the 55th Annual Meeting of the Association
  for Computational Linguistics (Volume 1: Long Papers)}, pages 1946--1958,
  Vancouver, Canada. Association for Computational Linguistics.

\bibitem[{Pires et~al.(2019)Pires, Schlinger, and
  Garrette}]{pires-etal-2019-multilingual}
Telmo Pires, Eva Schlinger, and Dan Garrette. 2019.
\newblock \href {https://doi.org/10.18653/v1/P19-1493} {How multilingual is
  multilingual {BERT}?}
\newblock In \emph{Proceedings of the 57th Annual Meeting of the Association
  for Computational Linguistics}, pages 4996--5001, Florence, Italy.
  Association for Computational Linguistics.

\bibitem[{Rahimi et~al.(2019)Rahimi, Li, and Cohn}]{rahimi-etal-2019-massively}
Afshin Rahimi, Yuan Li, and Trevor Cohn. 2019.
\newblock \href {https://doi.org/10.18653/v1/P19-1015} {Massively multilingual
  transfer for {NER}}.
\newblock In \emph{Proceedings of the 57th Annual Meeting of the Association
  for Computational Linguistics}, pages 151--164, Florence, Italy. Association
  for Computational Linguistics.

\bibitem[{Rapp(1995)}]{rapp-1995-identifying}
Reinhard Rapp. 1995.
\newblock \href {https://doi.org/10.3115/981658.981709} {Identifying word
  translations in non-parallel texts}.
\newblock In \emph{33rd Annual Meeting of the Association for Computational
  Linguistics}, pages 320--322, Cambridge, Massachusetts, USA. Association for
  Computational Linguistics.

\bibitem[{Ravi and Knight(2008)}]{ravi-knight-2008-attacking}
Sujith Ravi and Kevin Knight. 2008.
\newblock \href {https://www.aclweb.org/anthology/D08-1085} {Attacking
  decipherment problems optimally with low-order {N}-gram models}.
\newblock In \emph{Proceedings of the 2008 Conference on Empirical Methods in
  Natural Language Processing}, pages 812--819, Honolulu, Hawaii. Association
  for Computational Linguistics.

\bibitem[{Ravi and Knight(2011)}]{ravi-knight-2011-deciphering}
Sujith Ravi and Kevin Knight. 2011.
\newblock \href {https://www.aclweb.org/anthology/P11-1002} {Deciphering
  foreign language}.
\newblock In \emph{Proceedings of the 49th Annual Meeting of the Association
  for Computational Linguistics: Human Language Technologies}, pages 12--21,
  Portland, Oregon, USA. Association for Computational Linguistics.

\bibitem[{Ruder et~al.(2019)Ruder, Vulić, and Søgaard}]{Ruder_2019}
Sebastian Ruder, Ivan Vulić, and Anders Søgaard. 2019.
\newblock \href {https://doi.org/10.1613/jair.1.11640} {A survey of
  cross-lingual word embedding models}.
\newblock \emph{Journal of Artificial Intelligence Research}, 65:569–631.

\bibitem[{Shannon(1949)}]{shannon-1949-communication}
C.~E. Shannon. 1949.
\newblock \href {https://doi.org/10.1002/j.1538-7305.1949.tb00928.x}
  {Communication theory of secrecy systems}.
\newblock \emph{The Bell System Technical Journal}, 28(4):656--715.

\bibitem[{S{\o}gaard et~al.(2018)S{\o}gaard, Ruder, and
  Vuli{\'c}}]{sogaard-etal-2018-limitations}
Anders S{\o}gaard, Sebastian Ruder, and Ivan Vuli{\'c}. 2018.
\newblock \href {https://doi.org/10.18653/v1/P18-1072} {On the limitations of
  unsupervised bilingual dictionary induction}.
\newblock In \emph{Proceedings of the 56th Annual Meeting of the Association
  for Computational Linguistics (Volume 1: Long Papers)}, pages 778--788,
  Melbourne, Australia. Association for Computational Linguistics.

\bibitem[{Sérasset(2015)}]{serasset-gilles-2015-dbnary}
Gilles Sérasset. 2015.
\newblock \href {https://doi.org/10.3233/SW-140147} {Dbnary: Wiktionary as a
  lemon-based multilingual lexical resource in rdf}.
\newblock \emph{Semantic Web}, 6:355--361.

\bibitem[{Vincent et~al.(2008)Vincent, Larochelle, Bengio, and
  Manzagol}]{vincent-etal-2008-extracting}
Pascal Vincent, Hugo Larochelle, Yoshua Bengio, and Pierre-Antoine Manzagol.
  2008.
\newblock \href {https://doi.org/10.1145/1390156.1390294} {Extracting and
  composing robust features with denoising autoencoders}.
\newblock In \emph{ICML '08}, page 1096–1103, New York, NY, USA. Association
  for Computing Machinery.

\bibitem[{Vuli{\'c} and Moens(2013)}]{vulic-moens-2013-study}
Ivan Vuli{\'c} and Marie-Francine Moens. 2013.
\newblock \href {https://www.aclweb.org/anthology/D13-1168} {A study on
  bootstrapping bilingual vector spaces from non-parallel data (and nothing
  else)}.
\newblock In \emph{Proceedings of the 2013 Conference on Empirical Methods in
  Natural Language Processing}, pages 1613--1624, Seattle, Washington, USA.
  Association for Computational Linguistics.

\bibitem[{Vulic and Moens(2016)}]{Vulic:2016}
Ivan Vulic and Marie-Francine Moens. 2016.
\newblock \href {http://dl.acm.org/citation.cfm?id=3013558.3013583} {Bilingual
  distributed word representations from document-aligned comparable data}.
\newblock \emph{J. Artif. Int. Res.}, 55(1):953--994.

\bibitem[{Williams et~al.(2018)Williams, Nangia, and
  Bowman}]{williams-etal-2018-broad}
Adina Williams, Nikita Nangia, and Samuel Bowman. 2018.
\newblock \href {https://doi.org/10.18653/v1/N18-1101} {A broad-coverage
  challenge corpus for sentence understanding through inference}.
\newblock In \emph{Proceedings of the 2018 Conference of the North {A}merican
  Chapter of the Association for Computational Linguistics: Human Language
  Technologies, Volume 1 (Long Papers)}, pages 1112--1122, New Orleans,
  Louisiana. Association for Computational Linguistics.

\bibitem[{Wolf et~al.(2020)Wolf, Debut, Sanh, Chaumond, Delangue, Moi, Cistac,
  Rault, Louf, Funtowicz, Davison, Shleifer, von Platen, Ma, Jernite, Plu, Xu,
  Le~Scao, Gugger, Drame, Lhoest, and Rush}]{wolf-etal-2020-transformers}
Thomas Wolf, Lysandre Debut, Victor Sanh, Julien Chaumond, Clement Delangue,
  Anthony Moi, Pierric Cistac, Tim Rault, Remi Louf, Morgan Funtowicz, Joe
  Davison, Sam Shleifer, Patrick von Platen, Clara Ma, Yacine Jernite, Julien
  Plu, Canwen Xu, Teven Le~Scao, Sylvain Gugger, Mariama Drame, Quentin Lhoest,
  and Alexander Rush. 2020.
\newblock \href {https://doi.org/10.18653/v1/2020.emnlp-demos.6} {Transformers:
  State-of-the-art natural language processing}.
\newblock In \emph{Proceedings of the 2020 Conference on Empirical Methods in
  Natural Language Processing: System Demonstrations}, pages 38--45, Online.
  Association for Computational Linguistics.

\bibitem[{Wu and Dredze(2020{\natexlab{a}})}]{wu-dredze-2020-languages}
Shijie Wu and Mark Dredze. 2020{\natexlab{a}}.
\newblock \href {https://doi.org/10.18653/v1/2020.repl4nlp-1.16} {Are all
  languages created equal in multilingual {BERT}?}
\newblock In \emph{Proceedings of the 5th Workshop on Representation Learning
  for NLP}, pages 120--130, Online. Association for Computational Linguistics.

\bibitem[{Wu and Dredze(2020{\natexlab{b}})}]{wu-dredze-2020-explicit}
Shijie Wu and Mark Dredze. 2020{\natexlab{b}}.
\newblock \href {https://doi.org/10.18653/v1/2020.emnlp-main.362} {Do explicit
  alignments robustly improve multilingual encoders?}
\newblock In \emph{Proceedings of the 2020 Conference on Empirical Methods in
  Natural Language Processing (EMNLP)}, pages 4471--4482, Online. Association
  for Computational Linguistics.

\bibitem[{Yang et~al.(2019)Yang, Zhang, Tar, and
  Baldridge}]{yang-etal-2019-paws}
Yinfei Yang, Yuan Zhang, Chris Tar, and Jason Baldridge. 2019.
\newblock \href {https://doi.org/10.18653/v1/D19-1382} {{PAWS}-{X}: A
  cross-lingual adversarial dataset for paraphrase identification}.
\newblock In \emph{Proceedings of the 2019 Conference on Empirical Methods in
  Natural Language Processing and the 9th International Joint Conference on
  Natural Language Processing (EMNLP-IJCNLP)}, pages 3687--3692, Hong Kong,
  China. Association for Computational Linguistics.

\bibitem[{Zhang et~al.(2017)Zhang, Liu, Luan, and
  Sun}]{zhang-etal-2017-adversarial}
Meng Zhang, Yang Liu, Huanbo Luan, and Maosong Sun. 2017.
\newblock \href {https://doi.org/10.18653/v1/P17-1179} {Adversarial training
  for unsupervised bilingual lexicon induction}.
\newblock In \emph{Proceedings of the 55th Annual Meeting of the Association
  for Computational Linguistics (Volume 1: Long Papers)}, pages 1959--1970,
  Vancouver, Canada. Association for Computational Linguistics.

\bibitem[{Zhang et~al.(2019)Zhang, Baldridge, and He}]{zhang-etal-2019-paws}
Yuan Zhang, Jason Baldridge, and Luheng He. 2019.
\newblock \href {https://doi.org/10.18653/v1/N19-1131} {{PAWS}: Paraphrase
  adversaries from word scrambling}.
\newblock In \emph{Proceedings of the 2019 Conference of the North {A}merican
  Chapter of the Association for Computational Linguistics: Human Language
  Technologies, Volume 1 (Long and Short Papers)}, pages 1298--1308,
  Minneapolis, Minnesota. Association for Computational Linguistics.

\bibitem[{Zhang et~al.(2016)Zhang, Gaddy, Barzilay, and
  Jaakkola}]{zhang-etal-2016-ten}
Yuan Zhang, David Gaddy, Regina Barzilay, and Tommi Jaakkola. 2016.
\newblock \href {https://doi.org/10.18653/v1/N16-1156} {Ten pairs to tag {--}
  multilingual {POS} tagging via coarse mapping between embeddings}.
\newblock In \emph{Proceedings of the 2016 Conference of the North {A}merican
  Chapter of the Association for Computational Linguistics: Human Language
  Technologies}, pages 1307--1317, San Diego, California. Association for
  Computational Linguistics.

\bibitem[{Zweigenbaum et~al.(2018)Zweigenbaum, Sharoff, and
  Rapp}]{zweigenbaum-etal-2018-overview}
Pierre Zweigenbaum, Serge Sharoff, and Reinhard Rapp. 2018.
\newblock Overview of the third bucc shared task: Spotting parallel sentences
  in comparable corpora.
\newblock In \emph{Proceedings of 11th Workshop on Building and Using
  Comparable Corpora}, pages 39--42.

\end{thebibliography}

\appendix

\end{document}